\documentclass[letterpaper, 10 pt, conference]{ieeeconf}  
\usepackage{amsmath, amssymb}
\usepackage{algorithm}
\usepackage{algpseudocode}
\usepackage{graphicx}
\usepackage{caption}
\usepackage{subcaption}
\usepackage[hidelinks]{hyperref} 

\IEEEoverridecommandlockouts                              

\title{\LARGE \bf
RAY-TOLD: Ray-Based Latent Dynamics for \\ Dense Dynamic Obstacle Avoidance with TDMPC 
}

\author{Seungho Han$^{1}$, Seokju Lee$^{2}$, and Jeonguk Kang$^{3}$
\thanks{*This work was not supported by any organization \textit{(Corresponding Author: Jeonguk Kang)}}
\thanks{$^{1}$Seungho Han is with the School of Electrical Engineering, Hanyang University, Ansan 15588, South Korea. (email: {\tt\small seunghohan@hanyang.ac.kr})}%
\thanks{$^{2}$Seokju Lee is with the Mechatronics, Systems and Control Lab (MSC Lab), Department of Mechanical Engineering, Korea Advanced Institute of Science and Technology (KAIST), Yuseong-gu, Daejeon 34141, South Korea.
        (email: {\tt\small dltjrwn0322@kaist.ac.kr})}%
\thanks{$^{3}$Jeonguk Kang is with the Samsung Research, Samsung Electronics, Seoul, South Korea.
        (email: {\tt\small kju8765@gmail.com})}%
}

\begin{document}

\maketitle
\thispagestyle{empty}
\pagestyle{empty}

\begin{abstract}
    Dense, dynamic crowds pose a persistent challenge for autonomous mobile robots. Purely reactive planning methods, such as Model Predictive Path Integral (MPPI) control, often fail to escape local minima in complex scenarios due to their limited prediction horizon. To bridge this gap, we propose Ray-based Task-Oriented Latent Dynamics (RAY-TOLD), a hybrid control architecture that integrates obstacle information into latent dynamics and utilizes the robustness of physics-based MPPI with the long-horizon foresight of reinforcement learning. RAY-TOLD leverages a LiDAR-centric latent dynamics model to encode high-dimensional sensor data into a compact state representation, enabling the learning of a terminal value function and a policy prior. We introduce a policy mixture sampling strategy that augments the MPPI candidate population with trajectories derived from the learned policy, effectively guiding the planner towards the goal while maintaining kinematic feasibility. Extensive tests in a stochastic environment with high-density dynamic obstacles demonstrate that our method outperforms the MPPI baseline, reducing the collision rate. The results confirm that blending short-horizon physics-based rollouts with learned long-horizon intent significantly enhances navigation reliability and safety.
\end{abstract}

\section{Introduction}

\begin{figure}[t]
    \centering
    \includegraphics[width=0.95\linewidth]{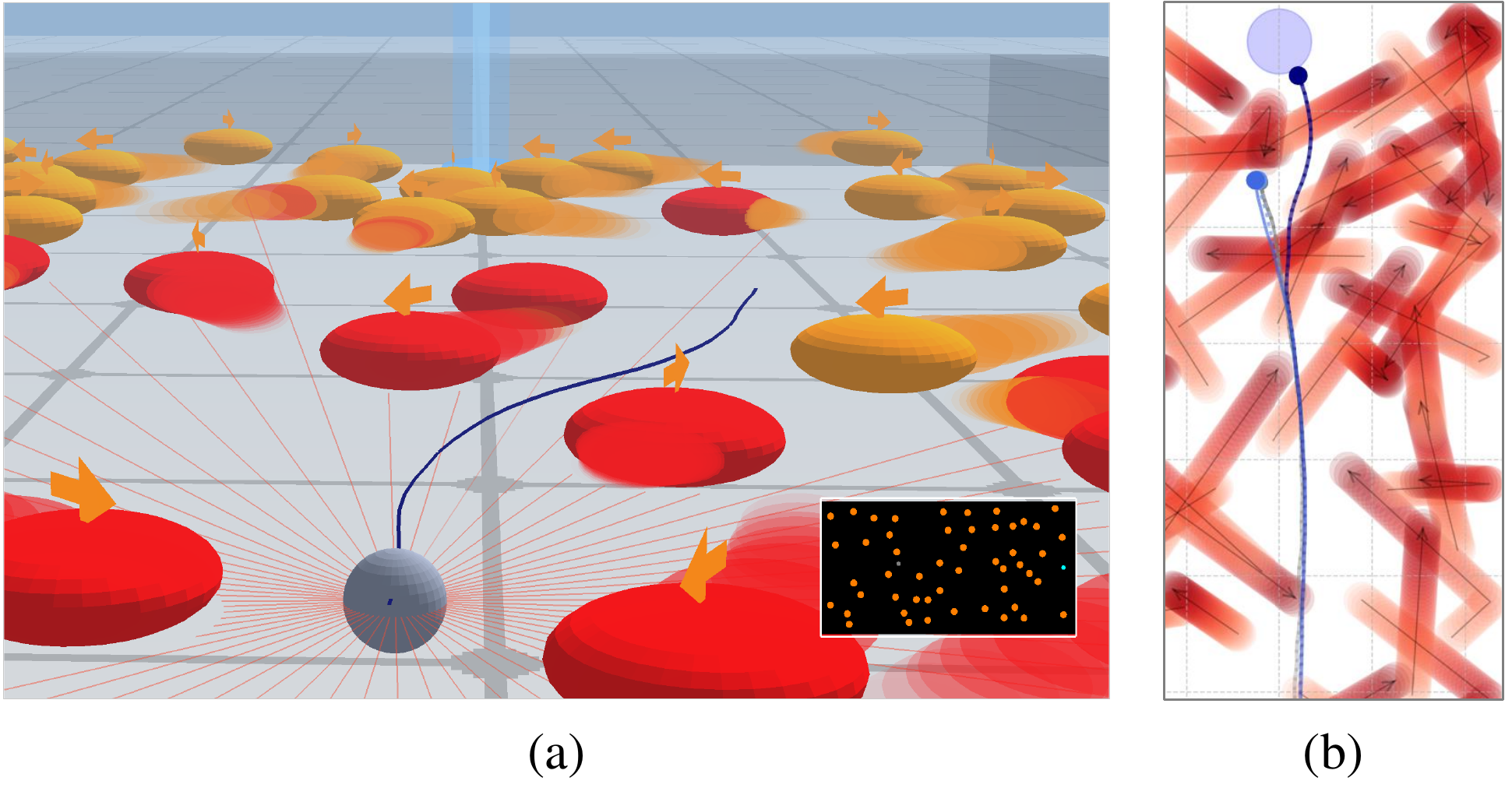}
    \caption{(a) An environment with dense dynamic obstacles and (b) the corresponding bird’s-eye view.}
    \label{fig:test}
\end{figure}

Navigating through dense crowds of dynamic obstacles (Fig. \ref{fig:test}) remains a formidable challenge for autonomous mobile robots. Real-world deployment in human-centric environments requires navigation systems capable of rapidly anticipating environmental changes, avoiding unpredictable dynamic agents, and strictly adhering to kinematic constraints~\cite{han2025dr}. While classical optimization-based methods provide reliable collision avoidance, the exponential growth of complexity in highly dynamic environments often necessitates a trade-off between computational efficiency and long-term foresight. 

Model Predictive Path Integral (MPPI) control has emerged as a powerful sampling-based method for handling non-linear vehicle dynamics in uncertain environments~\cite{crestaz2025td}. By leveraging parallel trajectory rollouts, MPPI achieves robust short-term collision avoidance and kinematic feasibility. However, purely reactive planners like standard MPPI suffer from a critical limitation: the computational cost of simulating forward dynamics curtails the feasible prediction horizon~\cite{wen2024collision}. Consequently, these methods lack long-term spatial awareness, frequently causing the robot to become trapped in local minima such as U-shaped obstacles~\cite{han2026fast} or dense crowd formations, where short-term safe paths fail to align with the global navigation goal. 

Reinforcement learning (RL), on the other hand, excels in consolidating long-horizon reasoning directly from environmental interactions. By learning a value function that approximates the infinite-horizon cost-to-go, RL provides agents with an intuitive foresight that helps avoid myopic decisions and local minima~\cite{lu2024mpc}. Furthermore, recent advancements in world models and latent dynamics have enabled agents to acquire structured and predictive representations of the environment, facilitating planning through learned internal models rather than relying solely on high-dimensional raw observations~\cite{shanks2025dreamernav, scannell2024iqrl}.

Despite these advances in model-based and representation-driven RL, many practical robotic systems still rely on end-to-end model-free RL policies due to their simplicity and scalability. However, such approaches often struggle to enforce hard safety constraints during learning and tend to suffer from inefficient or unsafe exploration, particularly in environments with complex or dynamic obstacles. These limitations pose significant challenges for the reliable deployment of RL policies in safety-critical physical robotic systems~\cite{romero2025actor}.

To bridge the gap between the constraint-satisfying robustness of physics-based MPC and the long-horizon foresight of learning-based models, this paper proposes \textbf{Ray-based Task-Oriented Latent Dynamics (RAY-TOLD)}. Built upon the principles of temporal-difference learning for model predictive control~\cite{hansen2022temporal, hansen2023td}, RAY-TOLD introduces a hybrid control architecture tailored for LiDAR-equipped autonomous robots navigating in dynamic crowds. Our approach encodes high-dimensional obstacle information from LiDAR scans into a low-dimensional latent representation. Within this latent space, a transition dynamics model, a terminal value function, and an action policy are jointly trained.

During online execution, RAY-TOLD resolves the local minima problem inherent to MPPI by introducing a policy mixture sampling strategy. Instead of relying entirely on unbiased Gaussian noise for trajectory generation, RAY-TOLD augments the MPPI candidate population with trajectories proposed by the learned policy prior. The sampled trajectories are then evaluated using the learned terminal value function, which encapsulates the long-term intent of the navigation task. This allows the MPPI planner to effectively look beyond its limited short-horizon rollout, securely guiding the robot around complex dynamic obstacles while guaranteeing kinematic feasibility.

The main contributions of this work are summarized as follows:
\begin{itemize}
    \item We present RAY-TOLD, a hybrid architecture that seamlessly integrates LiDAR-centric latent dynamics modeling with sampling-based MPPI to tackle navigation in dense dynamic environments.
    \item We introduce a policy mixture sampling technique coupled with a learned terminal value function, effectively eliminating the vulnerability of MPPI to local minima without requiring explicitly handcrafted heuristics.
    \item We conduct extensive tests in highly stochastic environments, demonstrating that blending short-horizon physics-based rollouts with long-horizon learned intent significantly reduces collision rates and outperforms standard MPPI baselines in safety and reliability.
\end{itemize}

The remainder of the paper is organized as follows. Section II reviews the related literature on MPC and RL-based collision avoidance. Section III presents the detailed methodology of RAL-TOLD, including the mathematical formulations. Section IV provides test validation of RAL-TOLD in environments with dense dynamic obstacles. Finally, Section V concludes the paper.

\section{Related Work}
MPC and MPPI are widely used since they explicitly handle system dynamics and constraints~\cite{wen2024collision}. To extend their prediction horizons, hierarchical approaches like PRM-RL~\cite{faust2018prm} pair probabilistic roadmaps with RL. Recent MPPI variants integrate perceptive dynamics for off-road tasks~\cite{roth2025learned} and distributed frameworks for scalable multi-robot coordination~\cite{zhang2025toward}. Additionally, TD-CD-MPPI~\cite{crestaz2025td} embeds temporal-difference learning into MPPI to approximate infinite-horizon costs. Building on these, RAY-TOLD utilizes MPPI as the core planner but explicitly steers the sampling using a LiDAR-trained task-oriented latent model.

Combining RL's long-term reasoning with MPC's safety has inspired various hybrid models. AC-MPC~\cite{romero2025actor} embeds a differentiable MPC within an RL actor, while DR-MPC~\cite{han2025dr} adds residual RL corrections to nominal MPC commands. MPC-inspired RL~\cite{lu2024mpc} parameterizes controllers like QP solvers, and Meta-RL frameworks use MPC to adapt dynamically to changing environments~\cite{shin2022infusing}. Unlike residual or differentiable combinations, our method injects RL guidance directly into MPPI's proposal distribution, ensuring all actions remain grounded in physics-based rollouts.

World models \cite{ha2018world} improve sample efficiency by enabling latent imagination. Architectures like DreamerNav~\cite{shanks2025dreamernav} merge depth and occupancy maps for indoor navigation, while iQRL~\cite{scannell2024iqrl} prevents representation collapse via implicit quantization. For more complex settings, DAWM~\cite{zhang2025integrating} uses Transformers to predict state increments at intersections, and Performer-MPC~\cite{xiao2022learning} applies real-time attention for dynamic spatial encoding. RAY-TOLD adapts these concepts specifically for sparse 2D LiDAR streams, crafting representations optimized for local motion prediction of dense crowds.

TD-MPC~\cite{hansen2022temporal} jointly learns latent dynamics and terminal values for fast planning. TD-MPC2~\cite{hansen2023td} scales this for multi-task continuous control, and D-MPC~\cite{zhou2024diffusion} incorporates diffusion models for multi-step action proposals. To mitigate value overestimation from out-of-distribution rollouts, TD-M(PC)$^2$~\cite{lin2025td} introduces policy constraints during value learning. For hierarchical reasoning, IQL-TD-MPC~\cite{chitnis2024iql} sets subgoals via Implicit Q-Learning, while QT-TDM~\cite{kotb2024qt} uses an autoregressive Q-Transformer over a sequence model. Furthermore, SOMBRL~\cite{sukhija2025sombrl} demonstrates scalable optimistic exploration in model-based RL. RAY-TOLD extends this TD-MPC lineage to dense robotic navigation using policy mixture sampling to overcome dynamic obstacle stochasticity.

\begin{figure*}[t]
    \centering
    \includegraphics[width=1\linewidth]{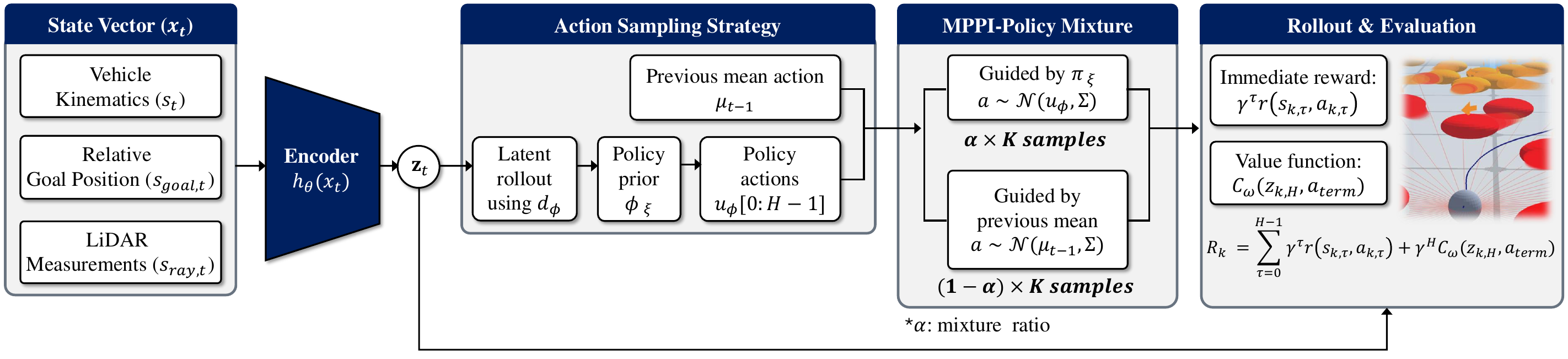}
    \caption{Overall architecture of the proposed RAY-TOLD model.}
    \label{fig:overall}
\end{figure*}

\section{Methodology}
\label{sec:methodology}
We present RAY-TOLD, a hybrid control framework that couples the robustness of physics-based model predictive control (MPC) with the long-horizon foresight of model-based reinforcement learning (MBRL). Our approach builds upon the conceptual foundation of task-oriented latent dynamics (TOLD)~\cite{hansen2023td}, developing a LiDAR-centric latent model that incorporates physics-based rollout for short-horizon interactions while leveraging a learned value function and policy prior to guide long-term decision-making.

\subsection{Problem Formulation}
We consider a discrete-time formulation of the optimal control problem. Let $s_t \in \mathcal{S}$ denote the state and $a_t \in \mathcal{A}$ the action at time $t$, evolving according to the system dynamics $s_{t+1} = f(s_t, a_t)$. The objective is to maximize the expected cumulative reward:
\begin{equation}
    J(\pi) = \mathbb{E}_{\pi} \left[ \sum_{t=0}^{\infty} \gamma^t r(s_t, a_t) \right],
\end{equation}
where $r(s_t, a_t)$ is the reward function and $\gamma \in [0, 1)$ is the discount factor. The $r(s_t, a_t)$ is designed to encourage efficient navigation while ensuring safety. It is composed of six terms:
\begin{equation}
    r = r_{dist} + r_{coll} + r_{side} + r_{vel} + r_{prog} + r_{goal}
\end{equation}

\begin{itemize}
    \item $r_{dist} = -\|\mathrm{p_t} - \mathrm{g}\|_2$: Negative Euclidean distance to the goal $\mathrm{g}$.
    \item $r_{coll} = -120 \cdot \mathbb{I}(d_{obs} < 0.1)$: Heavy penalty for collision, where $d_{obs}$ is the distance to the nearest obstacle surface.
    \item $r_{side} = -15 \cdot \exp(-4 \cdot d_{obs})$: Soft clearance penalty to encourage keeping a safe distance from obstacles.
    \item $r_{vel}$: Velocity incentive ($0.5v$) when far from the goal, switching to a braking penalty, i.e., $-v$ when within $2.0$ m.
    \item $r_{prog} = 5 \cdot (\mathbf{h}_t \cdot \mathbf{d}_{goal}) \cdot v$: Progress reward based on alignment between heading $\mathbf{h}_t$ and goal direction $\mathbf{d}_{goal}$.
    \item $r_{goal} = 300 \cdot \mathbb{I}(\|p_t - g\|_2 < 0.7)$: Sparse terminal reward for reaching the goal region.
\end{itemize}

\noindent\textbf{Notation:}
$\mathrm{p_t}\in\mathbb{R}^2$ denotes the ego position, $\mathrm{g}\in\mathbb{R}^2$ the goal position, and $\mathbb{I}(\cdot)$ the indicator function.
We define the goal-direction unit vector $\mathbf{d}_{goal} = (\mathbf{g}-\mathbf{p_t})\|\mathbf{g}-\mathbf{p_t}\|_2^{-1}$ and the heading unit vector $\mathbf{h}_t = [\cos\theta_t,\ \sin\theta_t]^\top$.
$d_{obs}$ is computed as the minimum LiDAR ray intersection distance to obstacle boundaries (i.e., truncated at the first hit), consistent with the occlusion-aware sensing model in Sec.~\ref{sec:verification}.


Our proposed architecture integrates two complementary prediction modalities, which address the primary limitations of the standard MPPI:
\begin{enumerate}
    \item \textbf{Policy-Guided Rollouts ($t \sim t{+}H{-}1$):} Unlike standard MPPI which relies solely on random noise perturbations, we perform $H$-step rollouts using a differentiable kinematic bicycle model where a fraction of the sampled actions are guided by a learned latent policy $\pi_\theta$. This mixture sampling explicitly biases the search space toward goal-directed, kinematically feasible trajectories while evaluating short-horizon collision outcomes.
    \item \textbf{Latent Terminal Value Estimation ($\ge t{+}H$):} Standard MPPI typically ignores the long-term consequences beyond the planning horizon $H$. Instead, we augment the finite-horizon objective with a learned terminal value. We encode the final state of the rollout into a compact latent representation $z_{t+H} = h_\theta(\mathbf{x}_{t+H})$ and use the learned value function $Q_\theta$ to approximate the expected return beyond the rollout horizon, effectively preventing the planner from being trapped in local minima.
\end{enumerate}

\begin{algorithm}[t]
\caption{Planning algorithms using RAY-TOLD model}
\label{alg:hybrid_mppi}
\begin{algorithmic}[1]
\State \textbf{Input:} Augmented state $\mathbf{x}_t$, Policy prior $\pi_\xi$, Value function $C_\omega$, Encoder $h_\theta$, Latent Dynamics $d_\phi$
\State \textbf{Hyperparameters:} Horizon $H$, Samples $K$, Mixture ratio $\alpha$, Temp $\lambda$, Iterations $M$
\State Initialize mean action sequence $\mu_{0:H-1} \leftarrow \text{shift}(\mu_{t-1})$
\State $u^\pi_{0:H-1} \leftarrow$ Rollout $\pi_\xi$ in latent space using $d_\phi$ from $h_\theta(\mathbf{x}_t)$
\For{$m = 1$ to $M$}
    \State Sample $\epsilon_{k, \tau} \sim \mathcal{N}(0, \Sigma)$ for $k=1 \dots K, \tau=0 \dots H-1$
    \For{$k = 1$ to $K$} 
        \If{$k \le \alpha K$} 
            \State $a_{k, 0:H-1} \leftarrow \text{clip}(u^\pi_{0:H-1} + \epsilon_{k, 0:H-1})$
        \Else
            \State $a_{k, 0:H-1} \leftarrow \text{clip}(\mu_{0:H-1} + \epsilon_{k, 0:H-1})$
        \EndIf
        \State $s_{k, 0} \leftarrow s_t, R_k \leftarrow 0$
        \For{$\tau = 0$ to $H-1$}
            \State $s_{k, \tau+1} \leftarrow f(s_{k, \tau}, a_{k, \tau})$ 
            \State $R_k \leftarrow R_k + \gamma^\tau r(s_{k, \tau}, a_{k, \tau})$ 
        \EndFor
        \State $z_{k, H} \leftarrow h_\theta(\mathbf{x}_{k, H})$
        \State $a_{\text{term}} \leftarrow \pi_\xi(z_{k, H})$
        \State $\Phi_{\text{term}} \leftarrow C_\omega(z_{k, H}, a_{\text{term}})$ 
        \State $R_k \leftarrow R_k + \gamma^H \Phi_{\text{term}}$
    \EndFor
    \State $\Omega \leftarrow \sum_{k=1}^K \exp(\frac{1}{\lambda} R_k), w_k \leftarrow \frac{1}{\Omega} \exp(\frac{1}{\lambda} R_k)$
    \State $\mu_\tau \leftarrow \sum_{k=1}^K w_k a_{k, \tau}$ for $\tau = 0 \dots H-1$
\EndFor
\State \textbf{Return:} mean action $\mu_0$
\end{algorithmic}
\end{algorithm}

\subsection{RAY-TOLD Model}
Parallel to the physics-based planner, we train a RAY-TOLD model to learn a compact latent representation $z_t$ and associated value dynamics, where the overall schematic is shown in Fig. \ref{fig:overall}. We define an augmented state vector $\mathbf{x}_t = [s_t, s^{goal}_t, s^{ray}_t]$, where $s_t$ is the vehicle kinematics, $s^{goal}_t$ is the relative goal position, and $s^{ray}_t$ comprises the $N_{rays}$ range measurements and obstacle velocities.

The encoder $h_\theta(\mathbf{x}_t)$ compresses this high-dimensional input into a low-dimensional latent state $z_t$. The individual components of the RAY-TOLD model are parameterized by separate neural networks:
\begin{align}
    z_t &= h_\theta(\mathbf{x}_t) & \text{(Encoder)} \\
    z_{t+1} &= d_\phi(z_t, a_t) & \text{(Latent Dynamics)} \\
    \hat{r}_t &= R_\psi(z_t, a_t) & \text{(Reward Predictor)} \\
    Q(z_t, a_t) &= C_\omega(z_t, a_t) & \text{(Value Function)} \\
    \hat{a}_t &\sim \pi_\xi(z_t) & \text{(Policy Prior)}
\end{align}
Here, the reward predictor $R_\psi$ and the value function $C_\omega$ serve complementary but distinct roles. The reward predictor $R_\psi(z_t, a_t) \approx r(s_t, a_t)$ is a neural network that estimates the \emph{immediate, single-step} reward directly from the latent state and action, without requiring access to the full physics engine. This is essential during training, where multi-step imagined rollouts are performed entirely within the learned latent space using the latent dynamics model $d_\phi$: since these rollouts do not pass through the true environment, the ground-truth reward function $r(s_t, a_t)$ is unavailable, and $R_\psi$ serves as its differentiable surrogate. The reward predictor thus enables end-to-end gradient-based optimization of the latent representation by providing the reward signal needed to train the value function and policy prior through temporal difference (TD) learning.

In contrast, the value function $C_\omega(z_t, a_t) \approx Q^\pi(s_t, a_t)$ serves as the critic, approximating the \emph{infinite-horizon discounted return} from a given latent state-action pair. During evaluation, $C_\omega$ is used exclusively at the terminal state of each MPPI rollout (i.e., at time $t{+}H$) to estimate the expected future return beyond the finite planning horizon. This terminal value estimation is the key mechanism by which RAY-TOLD overcomes the myopic limitation of standard MPPI.

\subsection{Model Training}
The RAY-TOLD model is trained end-to-end using trajectories collected in the environment. Following the TD-MPC2 framework~\cite{hansen2023td}, the training objective minimizes a combined loss function:
\begin{equation}
    \mathcal{L} = \lambda_1 \mathcal{L}_{reward} + \lambda_2 \mathcal{L}_{value} + \lambda_3 \mathcal{L}_{policy} + \lambda_4 \mathcal{L}_{latent}
\end{equation}
where $\mathcal{L}_{reward}$ trains $R_\psi$ to predict ground-truth single-step rewards, $\mathcal{L}_{value}$ trains $C_\omega$ via Temporal Difference (TD) learning, and $\mathcal{L}_{policy}$ optimizes the actor network $\pi_\xi$ to maximize the Q-value. Finally, $\mathcal{L}_{latent}$ enforces consistency between the state transitions predicted by the Latent Dynamics $d_\phi$ and the actual encoded future states, ensuring the latent space accurately captures the underlying transition physics.


Model Predictive Path Integral (MPPI) control is deployed for trajectory optimization. Standard MPPI draws action perturbations around the previous solution, which yields a unimodal proposal distribution. In dense, multimodal interaction scenarios, this often under-explores alternative homotopy classes and leads to local-minima.

Thus, we introduce a policy mixture sampling strategy, originally introduced in TD-MPC2~\cite{hansen2023td}. The MPPI population is augmented by seeding a fraction $\alpha$ of the samples with trajectories derived from the learned policy $\pi_\xi$. Specifically, let $u^\pi_{t:t+H-1}$ be the action sequence generated by recursively unrolling the policy $\pi_\xi$ using the Latent Dynamics network $d_\phi$. This allows generating multi-step future actions in the compact latent space without repeatedly invoking the expensive physics engine. The sampling distribution for the $k$-th candidate trajectory is given by:
\begin{equation}
    a^{(k)}_t \sim \begin{cases} 
    \mathcal{N}(a^\pi_t, \Sigma) & \text{if } k < \alpha N \\
    \mathcal{N}(a^{MPPI}_t, \Sigma) & \text{otherwise}
    \end{cases}
\end{equation}

\begin{table}[t]
    \caption{Hyperparameters for RAY-TOLD Training and Evaluation}
    \centering
    \resizebox{\columnwidth}{!}{%
    \begin{tabular}{|l|c|}
        \hline
        \textbf{Parameter} & \textbf{Value} \\
        \hline
        Horizon ($H$) & 30  \\
        Evaluation Steps & 300 \\
        MPPI samples ($K$) & 256\\
        MPPI optimization iterations ($M$) & 3  \\
        Softmax temperature for MPPI weighting ($\lambda$) & 1.0  \\
        Discount factor ($\gamma$) & 0.99 \\
        \hline
        Lidar Rays ($N_{\text{rays}}$) & 60 \\
        Obstacle Radius & 0.4 m\\
        Map Size & $20 \;\mathrm{m} \times 10\;\mathrm{m}$  \\
        \hline
        Batch Size & 256  \\
        Learning Rate & $1 \times 10^{-4}$  \\
        Replay Buffer Size & 50{,}000 \\
        \hline
    \end{tabular}}
    \label{tab:hyperparameters}
\end{table}

This mechanism injects global knowledge from the learned policy into the local optimization process of MPPI, effectively guiding the planner towards long-term objectives while retaining the reactivity of the physics-based model. The corresponding objective function is formulated by 

\begin{equation}
\resizebox{\columnwidth}{!}{$
J(a_{t:t+H-1}) = \mathbb{E}\!\left[\sum_{k=0}^{H-1}\gamma^k r(s_{t+k}, a_{t+k})
+ \gamma^H C_\omega\!\left(z_{t+H},\, \pi_\xi(z_{t+H})\right)\right],
$}
\end{equation}
where $z_{t+H} = h_\theta(\mathbf{x}_{t+H})$ is the latent representation of the terminal state, and $C_\omega$ evaluates its expected return following the policy prior $\pi_\xi$.

\begin{table*}[h]
\caption{Performance comparison. The proposed method achieves an improvement in success rate and reduction in collision rate compared to the MPPI baseline, demonstrating the benefit of RAY-TOLD model in complex scenarios.}
\label{tab:results}
\centering
\begin{tabular}{|l|c c|c c|c c|}
\hline
\textbf{Method} & \textbf{Success Rate} & \textbf{Improv.} & \textbf{Collision Rate} & \textbf{Improv.} & \textbf{Safety Margin} & \textbf{Improv.} \\
\hline
MPPI (Baseline) & 89.0\% & - & 11.0\% & - & 0.19m & - \\
\hline
RAY-TOLD ($\alpha{=}0$) & 89.0\% & 0.0\% & 11.0\% & 0.0\% & \textbf{0.17m} & \textbf{10.53\%} \\
RAY-TOLD ($\alpha{=}0.1$) & \textbf{94.0\%} & \textbf{5.62\%} & \textbf{6.0\%} & \textbf{45.45\%} & 0.18m & 5.26\% \\
RAY-TOLD ($\alpha{=}0.2$) & 91.0\% & 2.25\% & 9.0\% & 18.18\% & 0.18m & 5.26\% \\
\hline
\end{tabular}
\end{table*}

\begin{figure*}
    \centering
    \begin{subfigure}[b]{0.495\textwidth}
        \centering
        \includegraphics[width=\textwidth]{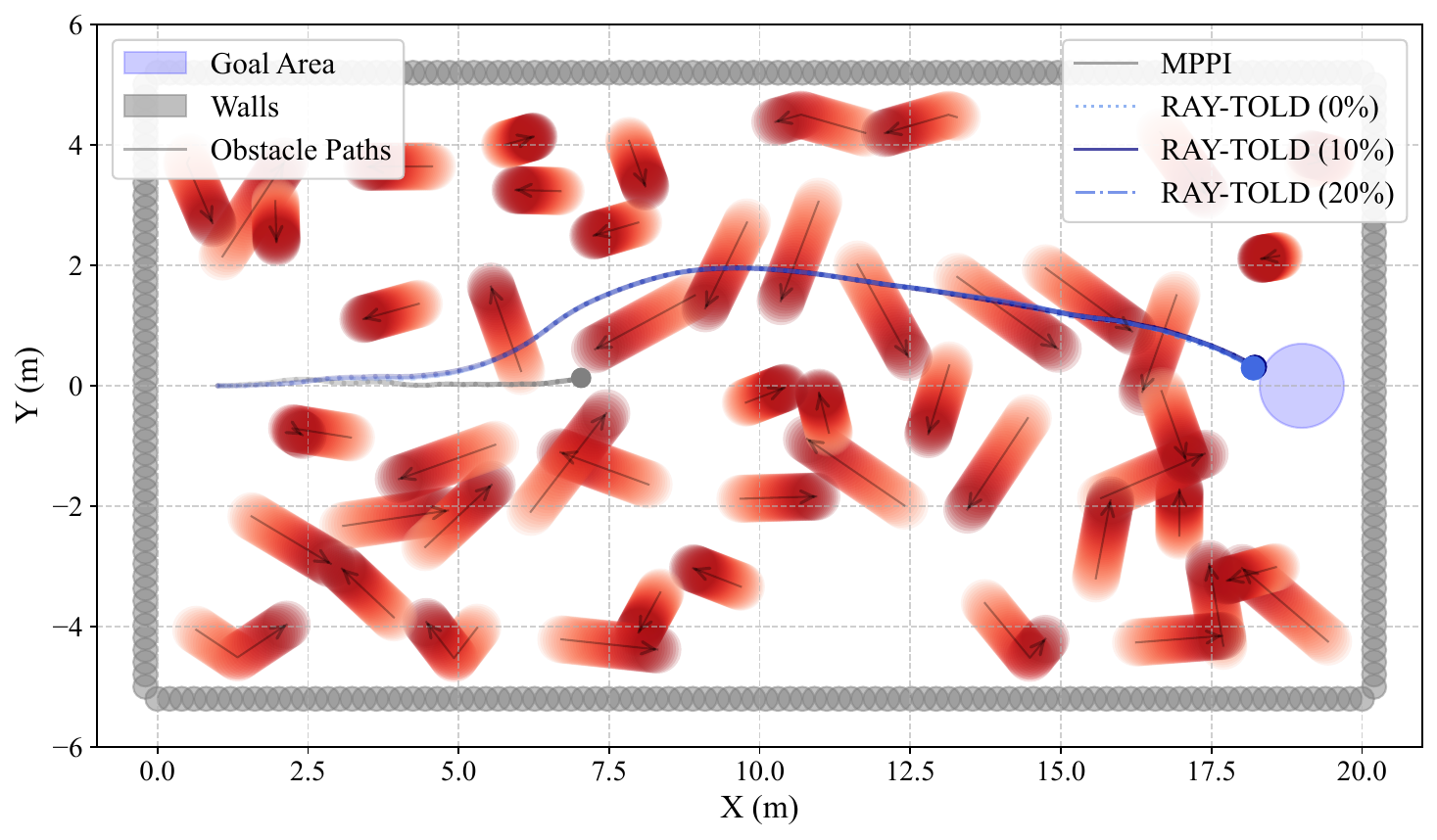}
        \caption{}
        \label{fig:traj_43}
    \end{subfigure}
    \hfill
    \begin{subfigure}[b]{0.495\textwidth}
        \centering
        \includegraphics[width=\textwidth]{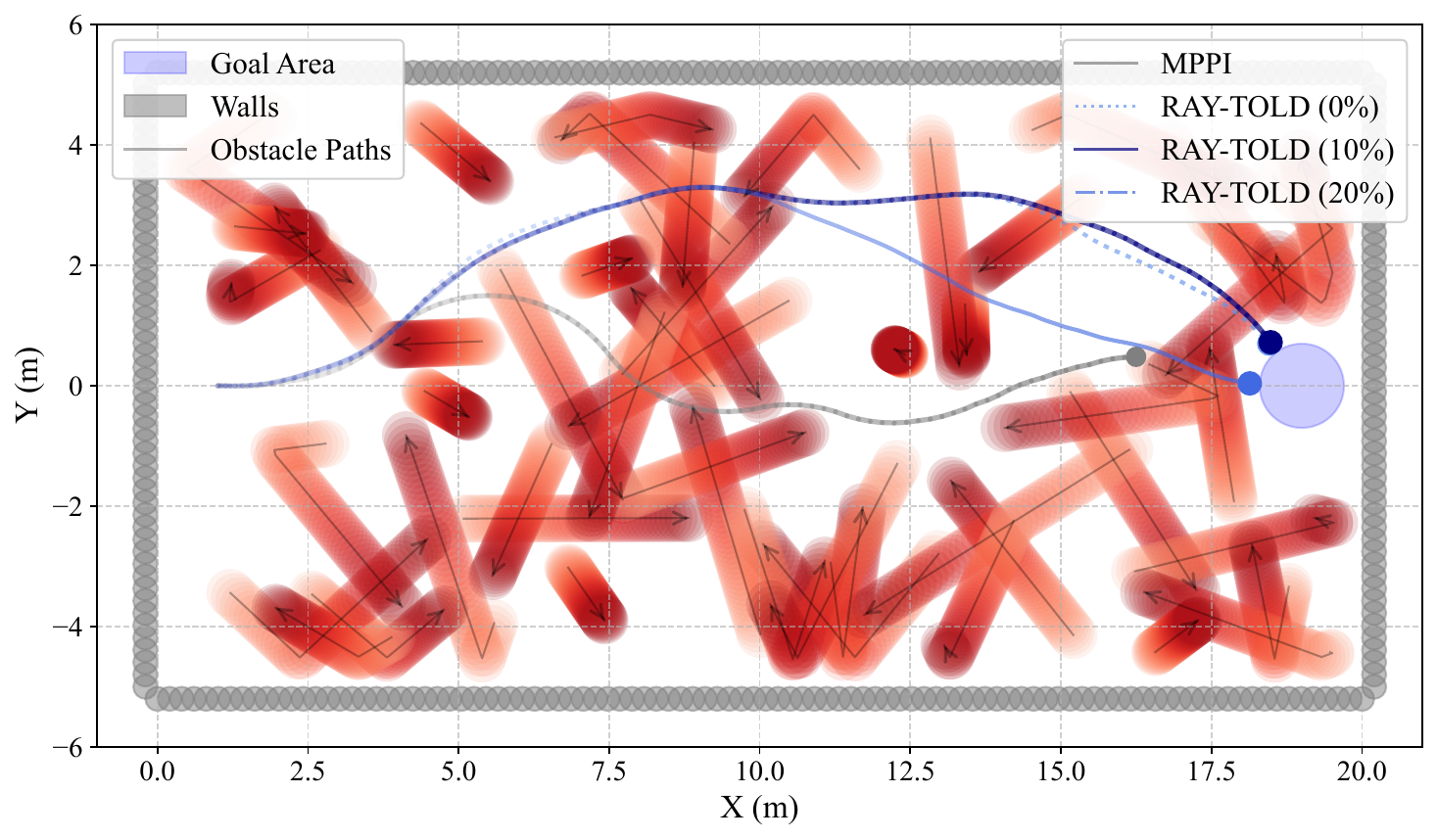}
        \caption{}
        \label{fig:traj_20}
    \end{subfigure}
    \vskip\baselineskip
    \begin{subfigure}[b]{0.495\textwidth}
        \centering
        \includegraphics[width=\textwidth]{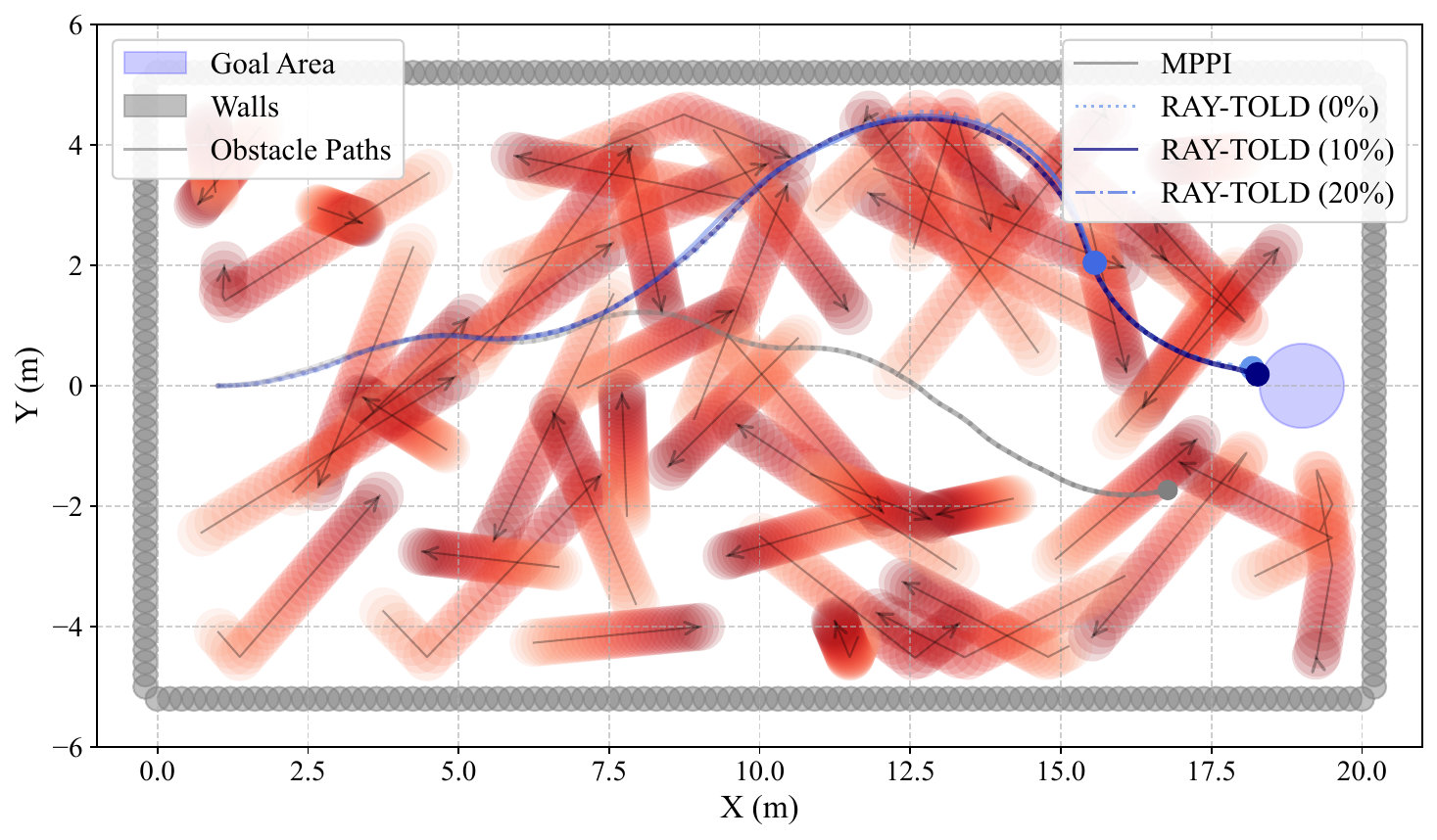}
        \caption{}
        \label{fig:traj_10}
    \end{subfigure}
    \hfill
    \begin{subfigure}[b]{0.495\textwidth}
        \centering
        \includegraphics[width=\textwidth]{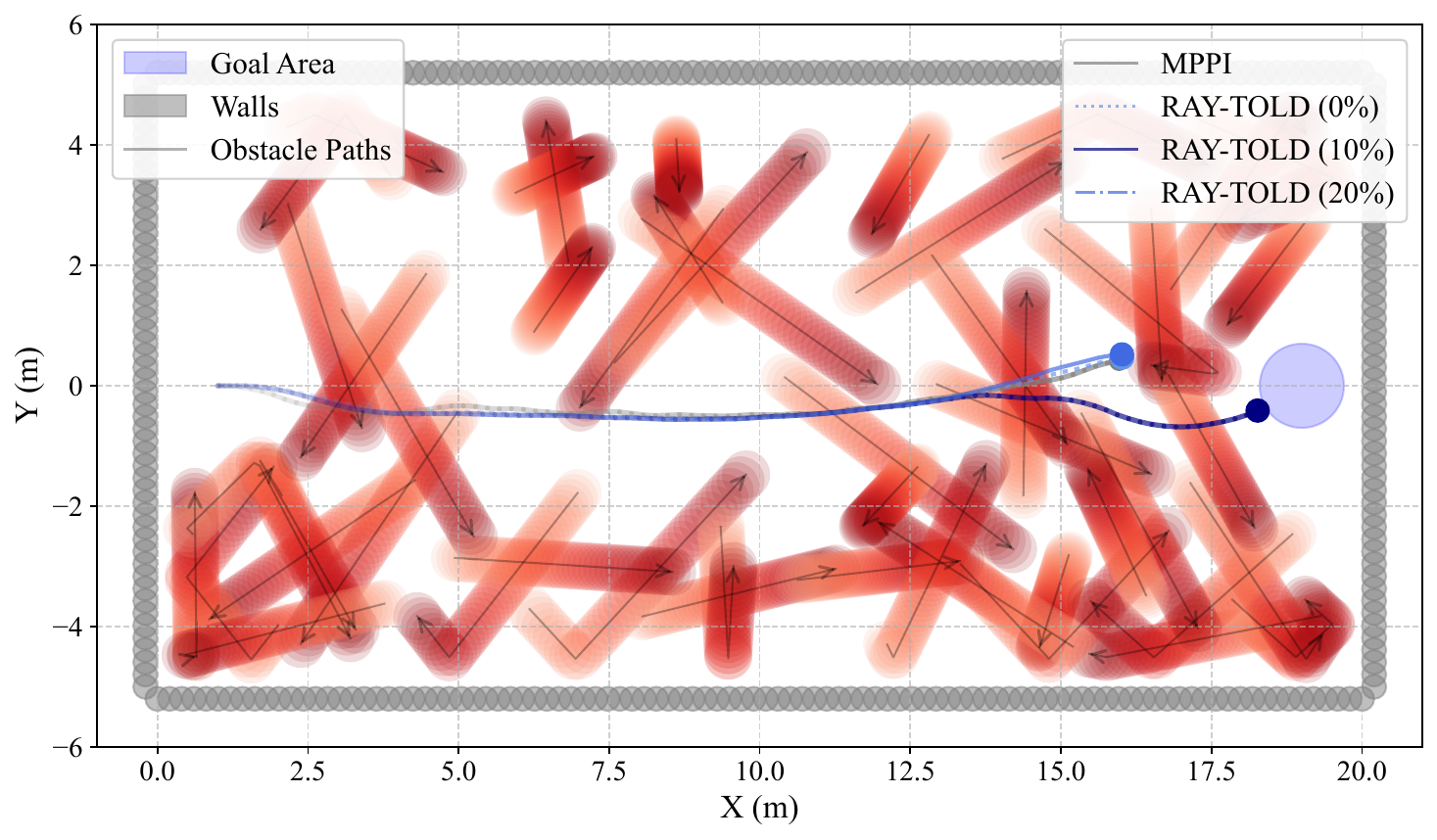}
        \caption{}
        \label{fig:traj_91}
    \end{subfigure}
    \caption{Four representative results from 100 test scenarios comparing trajectories in challenging dynamic environments. (a) Baseline failure, where MPPI fails to predict converging obstacles. (b) RAY-TOLD ($\alpha {=} 0.2$), where stronger policy guidance yields more efficient paths. (c) RAY-TOLD ($\alpha {=} 0.1,\;0.2$), which successfully anticipates obstacle reflections. (d) Only RAY-TOLD ($\alpha {=} 0.1$) avoids collision in this scenario.}
    \label{fig:qualitative_results}
\end{figure*}

\section{Verifications}
\label{sec:verification}

\subsection{Test Setup}
\noindent \textbf{Ego agent:} Our test platform is designed to rigorously evaluate the robustness of hybrid control in high-density, unpredictable environments. The primary evaluation arena is a $20\;\mathrm{m} \times 10 \;\mathrm{m}$ bounded physical space. The ego vehicle, modeled as a kinematic bicycle with a $2.5 \;\mathrm{m}$ wheelbase, is tasked with traversing from the coordinate $(1, 0)$ to the goal region centered at $(19, 0)$. To support the computational demands of this hybrid approach, we implemented a highly parallelized custom continuous-space physics simulation environment. This architecture supports simulating $N=256$ independent environments with $K=256$ rollout samples each, enabling real-time planning at 50 Hz without CPU-GPU data transfer bottlenecks. A vectorized benchmark is conducted over 100 episodes. Success is defined as reaching the goal without collision within the time limit.

The ego vehicle's state $s_t = [x, y, \theta, v]^T$ is updated at 10Hz. The action space is constrained to acceleration $a \in [-3, 3] m/s^2$ and steering angle $\delta \in [-45^\circ, 45^\circ]$. For environment sensing, the agent is equipped with a 360-degree LiDAR system providing $N_{\text{rays}}=60$ distance measurements. This sensor is occlusion-aware; rays are truncated upon the first intersection with an obstacle boundary, simulating realistic line-of-sight constraints.

\noindent \textbf{Obstacle:} As shown in Fig. \ref{fig:test}, where orange and red obstacles represent detected and undetected by ray beams, respectively, the environment contains a dynamic population of 40 to 60 obstacles, each with a radius of $0.4\;\mathrm{m}$. The obstacles are spawned at random positions with random velocity for each test scenario. Their behavior is governed by a modified Social Force Model (SFM). The movement of each obstacle $i$ is driven by three primary force components:
\begin{enumerate}
    \item Attractive Force: A force pulling the obstacle towards a randomly assigned local waypoint, ensuring purposeful motion.
    \item Boundary Repulsion: A repulsive potential that prevents obstacles from leaving the map, forcing them to turn back into the navigable space.
    \item Inter-Obstacle Repulsion: Repulsive forces between obstacles that prevent unrealistic overlaps and simulate basic collective avoidance behavior.
\end{enumerate}

A core design tenet of our benchmark is the prevention of environmental memorization. At the start of every training and evaluation episode, the obstacle positions, target waypoints, and velocities are completely randomized. This ensures that the global layout of safe zones changes continuously.

The inherent randomness of the SFM-driven trajectories means that even if the agent encounters a similar starting configuration, the temporal evolution of the obstacle field is non-deterministic from the agent's perspective. This high degree of stochasticity forces the Reinforcement Learning (RL) component to learn generalized features of the value landscape rather than memorizing specific evasion paths. By mastering this environment, the agent demonstrates that its terminal value function $Q_\theta$ and policy prior $\pi_\theta$ have captured the universal principles of dynamic avoidance and long-term goal seeking.

We perform a comparative analysis against a state-of-the-art Model Predictive Path Integral (MPPI) baseline:
\begin{enumerate}
    \item \textbf{MPPI (Baseline):} A purely physics-based reactive controller. It relies on a 3-second constant-velocity prediction model for dynamic obstacles. Crucially, it lacks a learned value function to evaluate the terminal state ($H$-step horizon), making it prone to local minima in dense clusters where the instantaneous gradient towards the goal is blocked.
    \item \textbf{RAY-TOLD ($\alpha{=}0$):} The proposed architecture without policy mixture sampling. It generates candidate trajectories using the same random noise as the MPPI baseline, but uniquely utilizes the learned value function $C_\omega$ to evaluate the long-term cost of the terminal latent state. This isolates the performance gain provided strictly by the terminal value estimation.
    \item \textbf{RAY-TOLD ($\alpha{=}0.1$):} The proposed configuration with moderate policy guidance.
    \item \textbf{RAY-TOLD ($\alpha{=}0.2$):} The proposed configuration with stronger policy guidance.
\end{enumerate}
We constrained the policy mixture ratio $\alpha$ to the range $[0,\, 0.2]$ to preserve the reactive, physics-based collision avoidance capabilities of MPPI for the majority of the sampling population, while injecting just enough learned prior to guide the planner out of local minima.

The detailed parameters are given in Table \ref{tab:hyperparameters}.

\subsection{Results}
We configure 100 different test scenarios in which the obstacles’ initial positions and velocities are randomly assigned for each scenario. The test results are listed in Table \ref{tab:results}, where the RAY-TOLD models are compared with the baseline method, i.e., MPPI. The evaluation metrics include success rate, collision rate, and the corresponding improvement for each metric. In addition, a safety margin is reported, defined as the closest distance between the ego agent and the obstacle, averaged over the entire test duration. Representative results comparing the ego trajectories are shown in Fig. \ref{fig:qualitative_results}.

According to Table \ref{tab:results}, the MPPI baseline demonstrates sufficient collision avoidance performance across all test scenarios, which involve highly dense dynamic obstacles resulting in intractable configurations. MPPI achieves an 89\% success rate out of 100 scenarios, with a safety margin of 0.19 m. Collisions occur due to the unpredictable movements of obstacles, such as abrupt changes in direction caused by reflections from other obstacles or pop-up obstacles hidden behind detected ones. This infeasibility of accurately modeling obstacle motion underlies the necessity of the proposed RAY-TOLD model, which leverages latent dynamics.

In practice, the RAY-TOLD models exhibit comprehensive improvements across all metrics. Specifically, RAY-TOLD ($\alpha {=} 0$) achieves an identical success rate to the baseline, while improving the safety margin, thereby validating its effectiveness in collision avoidance. Meanwhile, RAY-TOLD ($\alpha {=} 0.1,\;0.2$) demonstrates significant improvements in navigation reliability. RAY-TOLD ($\alpha {=} 0.1$) achieves the highest success rate of 94.0\%, surpassing both the MPPI baseline and the value-only variant ($\alpha {=} 0$) by +5.0\%. The collision rate is reduced by nearly half (from 11.0\% to 6.0\%), indicating that moderate policy guidance effectively helps the agent navigate tight clusters of dynamic obstacles, where purely reactive or value-based methods may fail. Interestingly, increasing the mixture ratio to 20\% results in a slight performance drop to 91.0\%, suggesting that over-reliance on the policy prior may introduce suboptimal biases in highly stochastic scenarios. Therefore, a balanced mixture ($\alpha {=} 0.1$) provides the optimal trade-off between long-horizon intent and short-horizon safety.

As shown in Fig. \ref{fig:qualitative_results}(a), a failure case of the algorithm is illustrated, highlighting its inability to account for the complex nonlinear dynamics of converging obstacles. Although MPPI relies on stochastic sampling to explore the control space, scenarios in which obstacles close in on the agent’s predicted path suffer from severe sample sparsity, where the density of valid, collision-free trajectories decreases drastically. Without an informed prior to guide the sampling process, the random exploration fails to identify feasible gaps between the converging obstacles.

Fig. \ref{fig:qualitative_results}(b) shows RAY-TOLD ($\alpha {=} 0.2$) produces stronger policy guidance leading to more efficient paths.
In contrast, it demonstrates that the proposed RAY-TOLD framework, with the policy guidance parameter $\alpha$ set to 0.2, generates highly efficient trajectories even in densely populated dynamic environments. By incorporating a learned policy as a strong prior for the MPPI planner, the algorithm effectively shifts the sampling mean toward promising regions of the action space. This policy-informed initialization significantly reduces the effective search space, enabling the agent to proactively identify and commit to optimal evasive maneuvers amid converging threats.

Fig. \ref{fig:qualitative_results}(c) highlights the advanced predictive capabilities of the RAY-TOLD framework, particularly its ability to anticipate obstacle reflections from environmental boundaries. In dynamic settings where obstacles bounce off walls or other constraints, traditional reactive planners often fail because they typically assume constant-velocity motion. However, by leveraging a latent-space world model integrated with informed policy priors, RAY-TOLD (with $\alpha {=} 0.1$ and $\alpha {=} 0.2$) effectively captures environmental transition dynamics and boundary interactions. This foresight enables the agent to execute preemptive maneuvers based on the obstacles’ future reflected trajectories, maintaining a safe distance even before a change in obstacle direction occurs.

In accordance with Fig. \ref{fig:qualitative_results}(d), only RAY-TOLD ($\alpha {=} 0.1$) avoids collision in this scenario.
It presents a critical corner case in which only the RAY-TOLD configuration with a specific guidance strength ($\alpha {=} 0.1$) successfully avoids collision. In this highly congested and complex scenario, the trade-off between exploiting the learned policy and exploring through the MPPI stochastic search becomes paramount. A higher $\alpha$ may overemphasize the policy prior, which can be slightly imprecise in extreme edge cases, while the baseline MPPI lacks sufficient guidance to discover the narrow escape route. In contrast, the $\alpha {=} 0.1$ setting provides an effective balance.

\begin{figure}[t]
    \centering
    \includegraphics[width=1\linewidth]{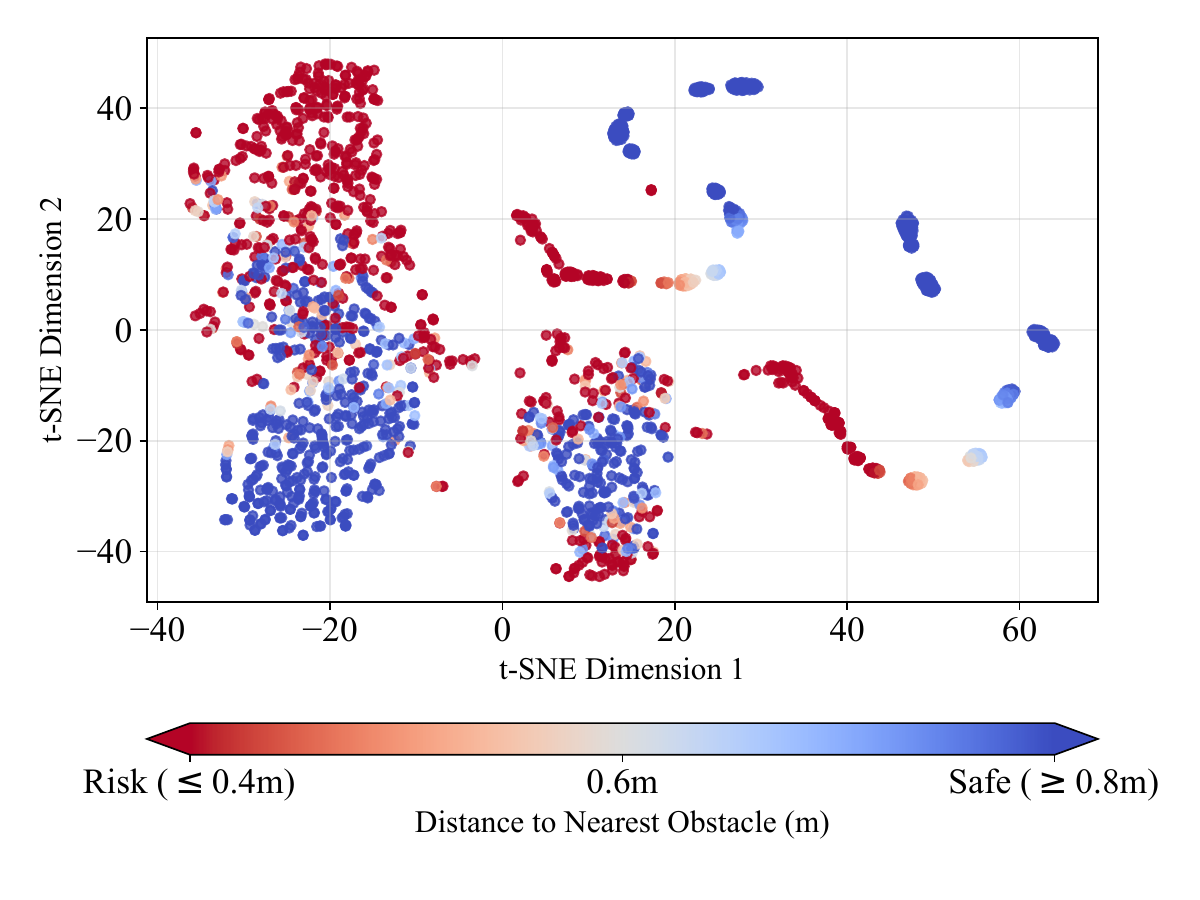}
    \caption{t-SNE visualization of the 128-dimensional latent space $z_t$, colored by the distance to the nearest obstacle. The smooth topological transition from dangerous (red) to safe (blue) states demonstrates that the RAY-TOLD learns a physically meaningful representation of collision risk rather than memorizing raw sensor data.}
    \label{fig:tsne}
\end{figure}

To further validate the representations learned by the RAY-TOLD architecture, we conducted a t-SNE dimensionality reduction analysis on the 128-dimensional latent state vectors $z_t$ generated sequentially during the evaluation rollouts.
As shown in Fig. \ref{fig:tsne}, the data points are colored according to the agent's distance to the nearest dynamic obstacle at that timestep. The visualization reveals a highly structured latent manifold: states corresponding to imminent collision risk (red) cluster distinctly away from safe, open-space states (blue), with a smooth topological transition between them.
This strongly implies that the encoder network $h_\theta$ is not merely memorizing the high-dimensional, 186-element raw observation vector, but is fundamentally abstracting the physical geometry and local interaction risks. Consequently, this contiguous and physically meaningful latent space provides a robust foundation for both the Value Function $C_\omega$ to forecast long-term safety and the policy prior $\pi_\xi$ to smoothly guide the MPPI sampling process.

\section{Conclusion}
While the baseline MPPI struggles with converging obstacles due to sample sparsity and the lack of an informed prior, RAY-TOLD mitigates these limitations by incorporating a learned policy into the sampling process. With stronger guidance ($\alpha {=} 0.2$), it generates efficient, goal-directed trajectories by focusing exploration on promising regions of the action space. The latent-space world model further enables anticipation of complex dynamics, such as obstacle reflections, supporting proactive and safe maneuvers. Notably, a balanced guidance strength ($\alpha {=} 0.1$) achieves the most robust performance by effectively trading off policy exploitation and stochastic exploration, highlighting the importance of integrating predictive modeling with tunable policy-guided search.

\section*{Acknowledgement}
This work was supported by the research fund of Hanyang University (HY-2025-3158)



\bibliographystyle{IEEEtran}

\end{document}